%% file: main.tex
\documentclass[10pt,twocolumn,letterpaper]{article}

\usepackage{cvpr}

\PassOptionsToPackage{numbers,  sort&compress}{natbib}
\usepackage{natbib}
\usepackage[dvipsnames]{xcolor}
\usepackage{times}
\usepackage{epsfig}
\usepackage{graphicx}
\usepackage{amsmath}
\usepackage{amssymb}

\usepackage{url}
\usepackage{booktabs}
\usepackage{amsfonts}
\usepackage{nicefrac}
\usepackage{microtype}
\usepackage{multirow}
\usepackage{algorithm}
\usepackage{algorithmic}
\renewcommand{\algorithmiccomment}[1]{\bgroup~~~{\color{tealblue}{\# #1}}\egroup}

\usepackage{wrapfig}
\usepackage{graphicx}
\usepackage{comment}
\usepackage{amsmath,amssymb}

\usepackage{multirow}
\usepackage{ctable}
\usepackage{pifont}

\input{math_commands}

\input{macros.tex}

\usepackage[multiple]{footmisc}
\usepackage{enumitem}
\usepackage{symbols}
\usepackage{mdframed}

\definecolor{mybrown}{rgb}{0.87058824, 0.56078431, 0.01960784}
\definecolor{myblue}{rgb}{0.3372549 , 0.70588235, 0.91372549}
\definecolor{mypurple}{rgb}{0.8, 0.47058824, 0.7372549 }
\definecolor{myorange}{rgb}{0.835, 0.368, 0}
\definecolor{mygreen}{rgb}{0.00784314, 0.61960784, 0.45098039}
\definecolor{mygt}{rgb}{0.0078125 , 0.57421875, 0.40625}
\definecolor{mysp}{rgb}{0.84765625, 0.515625  , 0.0234375}
\definecolor{mygr}{rgb}{0.9607,0.9607,0.9607}
\definecolor{myoo}{rgb}{0.992,0.9176,0.9019}
\usepackage{pifont}
\usepackage[T1]{fontenc}
\usepackage[utf8]{inputenc}
\usepackage{pmboxdraw}
\definecolor{mycitecolor}{HTML}{529C24}
\definecolor{mycitecolor1}{HTML}{668925}
\definecolor{mycitecolor2}{HTML}{E33E33}
\usepackage[pagebackref=true,breaklinks=true,colorlinks,bookmarks=false, %
]{hyperref}
\usepackage[frozencache,cachedir=.]{minted}
\def\confName{CVPR}
\def\confYear{2022}

\begin{document}

\title{Total Variation Optimization Layers for Computer Vision}

\author{Raymond A. Yeh\textsuperscript{$\dagger$} \;\;\; Yuan-Ting Hu \;\;\; Zhongzheng Ren\;\;\; Alexander G. Schwing\\
Toyota Technological Institute at Chicago\textsuperscript{$\dagger$} \;\;\; University of Illinois at Urbana-Champaign\\
{\tt\small yehr@ttic.edu, \{ythu2, zr5, aschwing\}@illinois.edu}
}

\maketitle

\begin{abstract}
\input{abs.tex}
\end{abstract}

\input{sec_intro}

\input{sec_rel}
\input{sec_app}

\input{sec_exp}

\input{sec_conc}

\clearpage
{\small
\bibliographystyle{ieee_fullname}
\bibliography{implicit_tv}
}

\end{document}

%% file: math_commands.tex
\usepackage{amsmath,amsfonts,bm}

\def\1{\bm{1}}

\def\sp{space}

\def\vd{{\bm{d}}}

\def\vs{{\bm{s}}}

\def\vu{{\bm{u}}}

\def\vx{{\bm{x}}}
\def\vy{{\bm{y}}}

\def\mD{{\bm{D}}}

\def\mH{{\bm{H}}}
\def\mI{{\bm{I}}}

\def\mL{{\bm{L}}}
\def\mM{{\bm{M}}}

\def\mP{{\bm{P}}}
\def\mQ{{\bm{Q}}}

\def\mX{{\bm{X}}}
\def\mY{{\bm{Y}}}
\def\mZ{{\bm{Z}}}

\DeclareMathAlphabet{\mathsfit}{\encodingdefault}{\sfdefault}{m}{sl}
\SetMathAlphabet{\mathsfit}{bold}{\encodingdefault}{\sfdefault}{bx}{n}
\newcommand{\tens}[1]{\bm{\mathsfit{#1}}}

\def\tX{{\tens{X}}}
\def\tY{{\tens{Y}}}

\def\gO{{\mathcal{O}}}

\def\gR{{\mathcal{R}}}
\def\gS{{\mathcal{S}}}

\def\sR{{\mathbb{R}}}

\DeclareMathOperator*{\argmin}{arg\,min}

%% file: macros.tex
\definecolor{darkred}{rgb}{0.7,0.1,0.1}
\definecolor{darkgreen}{rgb}{0.1,0.7,0.1}
\definecolor{cyan}{rgb}{0.7,0.0,0.7}
\definecolor{dblue}{rgb}{0.2,0.2,0.8}
\definecolor{maroon}{rgb}{0.76,.13,.28}
\definecolor{burntorange}{rgb}{0.81,.33,0}
\definecolor{tealblue}{rgb}{0.212,0.459, 0.533}

\definecolor{mypink}{rgb}{0.93359375, 0.62109375, 0.83984375}

\definecolor{pp}{rgb}{0.43921569, 0.18823529, 0.62745098}
\definecolor{rr}{rgb}{0.5254902 , 0.00784314, 0.12941176}
\definecolor{bb}{rgb}{0.09019608, 0.23529412, 0.37647059}
\definecolor{yy}{rgb}{0.49803922, 0.3372549 , 0.0}
\definecolor{gg}{rgb}{0.02352941, 0.3372549 , 0.17647059}

%% file: abs.tex
Optimization within a layer of a deep-net has emerged as a new direction for deep-net layer design. However, there are two main challenges when applying these layers to computer vision tasks: (a) which optimization problem within a layer is useful?; (b) how to ensure that computation within a layer remains efficient? To study question (a), in this work, we propose total variation (TV) minimization as a layer for computer vision. Motivated by the success of total variation in image processing, we hypothesize that TV as a layer provides useful inductive bias for deep-nets too. We study this hypothesis on five computer vision tasks: image classification, weakly supervised object localization, edge-preserving smoothing, edge detection, and image denoising, improving over existing baselines. To achieve these results we had to address question (b): we developed a GPU-based projected-Newton method which is $37\times$ faster than existing solutions.

%% file: sec_intro.tex
\vspace{-0.5cm}
\section{Introduction}\label{sec:intro}
Optimization within a deep-net layer has emerged as a promising direction to designing building blocks of deep-nets~\cite{
Amos_Kolter_2017,
Agrawal_Amos_Barratt_Boyd_Diamond_Kolter_2019,
Gould_Hartley_Campbell_2021}. 
For this, optimization problems are viewed as a differentiable function, mapping its input to its exact solution. The derivative of this mapping can be computed via implicit differentiation. Combined, this provides all the ingredients for a deep-net ``layer.'' 

Designing effective layers for deep-nets is crucial for the success of deep learning. For example, convolution~\cite{Fukushima_1980, lecun1999object}, recurrence~\cite{rumelhart1986learning,hochreiter1997long}, normalization~\cite{ioffe2015batch, wu2018group}, attention~\cite{vaswani2017attention} layers and other specialized layers~\cite{scarselli2008graph,kipf2016semi,liu2017video,yeh2019chirality} are the fundamental building-blocks of modern computer vision models.
Recently, optimization layers, \eg, OptNet~\cite{Amos_Kolter_2017}, have also found applications in
reinforcement learning~\cite{Amos_Jimenez_Sacks_Boots_Kolter_2018},
logical reasoning~\cite{Wang_Donti_Wilder_Kolter_2019}, 
hyperparamter tuning~\cite{ren2020not,Barratt_Boyd_2021},
scene-flow estimation~\cite{Teed_Deng_2021}, and
graph-matching~\cite{Rolinek_Swoboda_Zietlow_Paulus_Musil_Martius_2020},
providing useful inductive biases for these tasks. 
Despite these successes, optimization as a layer has not been as widely adopted in computer vision 
because of two unanswered questions: (a) which optimization problem is useful?; (b) how to efficiently solve for the exact solution of the optimization problem if the input is reasonably high-dimensional?

In this work, we propose and study Total Variation (TV)~\cite{Rudin_Osher_Fatemi_1992} minimization as a layer within a deep-net for computer vision, specifically, the TV proximity operator. 
We are motivated by the fact that TV has had numerous successes in computer vision, incorporating the prior knowledge that images are piece-wise constant. Notably, TV has been used as a regularizer in applications such as image denoising~\cite{Chambolle_2004}, super-resolution~\cite{Marquina_Osher_2008}, stylization~\cite{Johnson_Alahi_Fei-Fei_2016}, and blind deconvolution~\cite{Chan_Wong_1998} to name a few. Because of these successes, we hypothesize that TV as a layer would be an effective building-block in deep-nets, enforcing piece-wise properties in an end-to-end manner.%

However, existing solutions~\cite{Agrawal_Amos_Barratt_Boyd_Diamond_Kolter_2019, Barbero_Sra_2018} which can support TV as a deep-net layer are limited. For example, CVXPYLayers~\cite{Agrawal_Amos_Barratt_Boyd_Diamond_Kolter_2019} supports back-propagation through disciplined convex programs. However, CVXPYLayers uses a generic solver and lacks GPU support.
While specialized solvers~\cite{Barbero_Sra_2018,jimenez2011fast} for TV minimization exists, they also lack GPU and batching support. Hence, to meaningfully study TV as a layer at the scale of a computer vision task, we need a fast GPU implementation. To achieve this goal, we developed a fast GPU TV solver with custom CUDA kernels. For the first time, this enables use of TV as a layer across computer vision tasks. Our implementation is $1770\times$ faster than a generic solver and $37\times$ faster than a specialized TV solver. 

With this fast implementation, we study the hypothesis of TV as a layer on five tasks, spanning from high-level to low-level computer vision: 
classification, object localization, edge detection, edge-aware smoothing, and image denoising. We incorporate TV layers into existing deep-nets, \eg, ResNet and VGGNet, and found them to improve results. 

{\noindent \bf Our Contributions:}
\begin{itemize}[leftmargin=0.6cm]
\vspace{-0.2cm}
\itemsep0em 
\item We propose total variation as a layer for use as a building block in deep-nets for computer vision tasks.
\item We develop a fast GPU-based TV solver. It significantly reduces training and inference time, allowing a TV layer to be incorporated into classic deep-nets. The implementation is publicly available.\footnote{\url{github.com/raymondyeh07/tv_layers_for_cv}}
\item We demonstrate efficacy and practicality of TV layers by evaluating on five different computer vision tasks.
\end{itemize}

%% file: sec_rel.tex
\section{Related Work}\label{sec:rel}
In the following we briefly discuss optimization within deep-net layers, the use of total variation (TV) in computer vision and existing TV solvers.

{\bf \noindent Optimization as a Layer.} 
Optimization is a crucial component in classical statistical inference, \ie, within the ``forward pass'' of deep-nets, of machine learning models~\cite{Domke_2012}. Earlier works in structure prediction and graphical models~\cite{NIPS2003_878d5691, tsochantaridis2005large,joachims2009cutting,yeh2017interpretable}  rely on the output of an optimization program to make a prediction. End-to-end approaches have also been developed~\cite{belanger2016structured,GraberNIPS2018,GraberNeurIPS2019}.

More recently, optimization  has been viewed as a layer in deep-nets.~\citet{Amos_Kolter_2017} propose to integrate quadratic programming into deep-nets. Other optimization programs have also been considered, \eg, cone programs~\cite{Agrawal_Amos_Barratt_Boyd_Diamond_Kolter_2019}
and integer programs~\cite{Pogancic_Paulus_Musil_Martius_Rolinek_2020}. 
This perspective of optimization as a layer has also led to new optimization-based deep-net architectures.~\citet{Bai_Kolter_Koltun_2019,Bai_Koltun_Kolter_2020} propose deep equilibrium models which encapsulate all the layers into a root-finding problem. Optimization as a layer has also been explored in optimizing rank metrics~\cite{Rolinek_Musil_Paulus_Vlastelica_Michaelis_Martius_2020}
and graph matching~\cite{Rolinek_Swoboda_Zietlow_Paulus_Musil_Martius_2020}.
Different from these works, we propose and explore optimization of TV as a layer on computer vision tasks.

{\bf \noindent Total Variation in Computer Vision.}
Total variation, proposed by~\citet{Rudin_Osher_Fatemi_1992},  has been applied in various computer vision applications, including, denoising~\cite{Chambolle_2004, Beck_Teboulle_2009, Osher_Burger_Goldfarb_Xu_Yin_2005}
deconvolution~\cite{Chan_Wong_1998, Perrone_Favaro_2014},
deblurring~\cite{Beck_Teboulle_2009},
inpainting~\cite{Shen_Chan_2002, Afonso_Bioucas_Dias_Figueiredo_2010,yeh2017semantic},
superresolution~\cite{Marquina_Osher_2008},
structure-texture decomposition~\cite{Aujol_Gilboa_Chan_Osher_2006},
and segmentation~\cite{Donoser_Urschler_Hirzer_Bischof_2009}.
Total variation has also been applied in deep-learning as a loss function for visualizing deep-net features~\cite{Mahendran_Vedaldi_2015}, for style transfer~\cite{Johnson_Alahi_Fei-Fei_2016}, and for image synthesis~\cite{Zhang_Song_Qi_2017}. These prior works explored TV regularized problems. Differently, we study how to incorporate TV as a layer into end-to-end trained deep-nets. 

{\bf \noindent Solvers for TV Regularized Problems.}
A common approach to solving TV Regularized problems is 
Proximal Gradient Descent (PGD)~\cite{rockafellar1976monotone}, which requires computation of the TV proximity operator. Methods for solving the TV proximity operator include the taut string algorithm~\cite{condat2013direct}, Newton-type methods~\cite{jimenez2011fast}, the Iterative-Shrinkage-Thresholding Algorithm (ISTA)~\cite{combettes2005signal}, and its fast counterpart FISTA~\cite{Beck_Teboulle_2009}, \etc.
Besides these optimization algorithms, deep-nets have also been used to ``approximate'' solutions (via unrolling) of TV-regularized problems, \eg, 
Learned ISTA (LISTA)~\cite{Gregor_LeCun_2010} and variants~\cite{sprechmann2012learning},
Learned AMP~\cite{borgerding2017amp}, and Learned PGD (LPGD)~\cite{cherkaoui2020learning}.  
These works are interested in using deep-nets to solve an optimization problem, \ie, learning  to predict the solution. Different from these works, we are interested in how to use an optimization problem (as a layer) to incorporate inductive biases into deep-nets.

%% file: sec_app.tex
\begin{figure}
\centering
\setlength{\tabcolsep}{1pt}
\small
\begin{tabular}{ccc}
\includegraphics[height=0.325\linewidth]{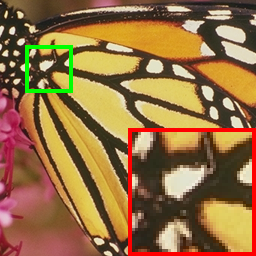} & 
\includegraphics[height=0.325\linewidth]{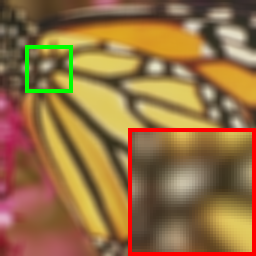} & 
\includegraphics[height=0.325\linewidth]{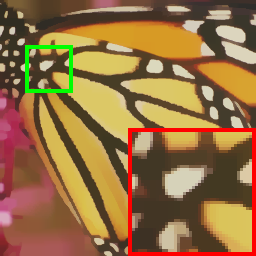}\\
Input & Conv.\ (Low-Pass) & Total Variation\\
\end{tabular}
\vspace{-0.13cm}
\caption{Illustration of convolution and TV proximity operator for image smoothing. In contrast to a convolution, \ie, a low-pass filter, TV is capable of preserving the edges during smoothing.}
\label{fig:tv_ill}
\vspace{-0.25cm}
\end{figure}

\section{Approach}\label{sec:app}
Our goal is to incorporate total variation (TV) minimization as a layer into deep-nets. %
Motivated by the success of TV in classical image restoration, we hypothesize that TV as a layer incorporated into deep-nets is a useful building-block for computer visions tasks. It  provides an additional selection of inductive bias over existing layers. Concretely, the input/output dependencies of a TV operation are not achievable by a single convolution layer, as the TV operation is not a linear system. Consider as an example image smoothing: TV can preserve the edges, while  a convolution (low-pass filter) blurs the edges as illustrated in~\figref{fig:tv_ill}. To incorporate this form of inductive bias into deep-nets we develop the differentiable TV layer.

\subsection{Differentiable Total Variation Layer}
The main component of our TV layer is the proximity operator. For 1D  input $\vx \in \sR^{N}$, TV is defined as
\be\label{eq:prox_tv1d}
\text{Prox}_{\text{TV}}^{1D}(\vx, \lambda)=\argmin_\vy \frac{1}{2}\norm{\vy-\vx}_2^2 + \lambda \norm{\mD_N\vy}_1,
\ee
where $\norm{\mD_N\vy}_1 = \sum_{n} |x_{n+1}-x_n|$ and $\lambda \geq 0$. The differencing matrix $\mD_N$ contains minus ones on its diagonal and an off-diagonal of ones,  capturing the gradients of $\vy$. 

Similarly, the 2D TV (anisotropic) proximity operation for a 2D input $\mX$ and output  $\mY \in \sR^{M \times N}$ is defined as:
\bea
&\text{Prox}_{\text{TV}}^{2D}(\mX, \lambda) = \argmin_\mY \frac{1}{2}\norm{\mY-\mX}_F^2\\
&+ \nonumber \lambda \left(\sum_m \norm{\mD_M \mY_{\tt row(m)}}_1 + \sum_n \norm{\mD_N \mY_{\tt col(n)}}_1\right),
\eea
where $\mD_M$ and $\mD_N$ denote the corresponding row and column differencing matrix.
We incorporate this TV proximity operator as a layer into deep-nets and refer to it as the \emph{differentiable total variation layer}.

{\noindent \bf Differentiable TV Layer.}
Given an input feature map $\tX \in \sR^{C\times H \times W}$ the TV layer outputs a tensor $\tY$ of the same size. This layer computes  the TV-proximity operator  independently on each of the channels.  The trainable parameters of this layer are $\tilde{\bm\lambda} \in \sR^C$ which are used in a SoftPlus~\cite{dugas2001incorporating} non-linearity to guarantee that $\bm\lambda$ contains positive numbers which are passed to the proximity operator. The forward operation is hence summarized as: $\forall c\in\{1, \dots, C\}$
\bea
\tY_{c} = \text{Prox}_{\text{TV}}^{2D}(\tX_c, \bm\lambda_c)\;\;\text{where}\;\;\bm\lambda_c = \text{SoftPlus}(\tilde{\bm\lambda}_c).
\eea
This layer performs smoothing while preserving edges. Note, depending on the desired spatial mode, the layer can also process the rows/columns independently, \ie, a 1D TV-proximity operator per row or column.

In addition, we further extend the capability of this layer, by designing a ``sharpening'' mode. Inspired by image sharpening techniques, we compute the difference between the input and the smoothed TV output. This difference is added back to the original image to perform ``sharpening,'' \ie, $\forall c$
\bea
\tY_{c} &=& 2\tX_c - \text{Prox}_{\text{TV}}^{2D}(\tX_c, \bm\lambda_c).
\eea
The overall pseudo-code of this layer is shown in~\figref{fig:pseudo_code}.

\input{figures/fig_pseudo_code}
{\noindent \bf Trainable $\bm\lambda$ in TV Layer.}
Note, a TV Layer with trainable $\bm\lambda$  increases a model's capacity. When $\bm\lambda=0$, this layer is an identity function for both smoothing and sharpening  as
\bea
 \vx = \argmin_\vy \frac{1}{2}\norm{\vy-\vx}_2^2 + 0,
\eea
and as $2\vx-\vx=\vx$ in the sharpening mode. Hence, a network with TV layers and $\bm\lambda=0$ is equivalent to a model without this layer. The network can hence learn to ``turn-off'' this layer if this improves results, avoiding the need to hand-tune $\bm\lambda$ when adding the TV layer to a new deep-net architecture.
We will now discuss  details on how to develop a fast implementation of this TV layer.

\subsection{Efficient Implementation}\label{sec:prox_tv_imp}
Existing packages that support the TV proximity operator are either  generic  or lack GPU support. For example, CVXPYLayers uses a generic solver,
which is relatively slow compared to specialized TV solvers. In contrast, efficient solvers are fast on the CPU,~\eg, the ProxTV toolbox~\cite{jimenez2011fast,Barbero_Sra_2018}, however CPU implementations are not suitable for integration with deep-nets when all other operations occur on the GPU, as memory transfer between GPU and CPU is needed at each layer, making training and inference slow. 

To address these short-comings, we develop a GPU-based Projected-Newton~\cite{bertsekas1982projected} method for solving TV by writing custom CUDA kernels. We carefully consider  the structure of the TV problem. These custom CUDA kernels are wrapped into PyTorch~\cite{pytorch_2019} and can be conveniently called through Python. We will discuss the forward and backward operation for 1D input next.

{\bf \noindent Forward Operation.} 
Projected-Newton solves the TV-proximity problem (\equref{eq:prox_tv1d}) via its dual:
\bea\label{eq:dual}
\max_{\vu} \underbrace{-\frac{1}{2} \norm{\mD_N^{\intercal}\vu}_2^2 + \vu^{\intercal}\mD_N\vx}_{\phi(\vu)}~~~\text{s.t.}~\norm{\vu}_{\infty} < \lambda.
\eea
Specifically, projected-Newton iteratively solves a local quadratic approximation of the objective before performing an update step with a suitable step-size, \eg, following the Armijo rule. Given~\equref{eq:dual}, the quadratic approximation boils down to solving the linear system
\bea\label{eq:solve_hess}
\mH_{\cal S} \vd_{\cal S} = \nabla\phi(\vu)_{\cal S},
\eea
for $\vd_{\cal S}$ which denotes the dual variable update direction. Here, $\mH=\nabla^2\phi(\vu)$ refers to the Hessian matrix,
$\gS$ refers to the set of indices of active variables and the subscript denotes selecting the rows/columns to form a system based on a subset of the variables.

At a glance, solving~\equref{eq:solve_hess} seems expensive. However, note that  $\mH=\mD_N\mD_N^{\intercal}$ is a tridiagonal and symmetric matrix. Tridiagonal systems can be solved efficiently by first computing a Cholesky factorization and then solving with backward substitution. Both operations can be performed in linear time, \ie, ${\gO}(N)$~\cite{alma99418280712205899}. Without exploiting this structure,   general Gaussian elimination has ${\gO}(N^3)$ complexity. 

Unfortunately, specialized routines for sub-indexing and solving tridiagonal systems are neither supported by cuBLAS nor by ATen. To enable efficient computation with batching support on the GPU, we implemented 21 custom CUDA kernels. These CUDA operations are integrated with PyTorch for ease of use. 
We note that these implementations are necessary due to the special structures of the matrices. To give an example, a $N \times N$ tridiagonal matrix can be efficiently stored in a $3 \times N$ matrix, storing the diagonal and two off-diagonals. However, this indexing scheme needs to be supported and is not readily available in existing packages. 

Next, to select a suitable step-size for direction $\vd_{\cal S}$, we use the quadratic interpolation backtracking strategy~\cite{alma99954904124605899}. We have also implemented a parallelized search strategy, which considers multiple step-sizes of halving intervals in parallel. In practice, we found   backtracking to be more efficient as it only iterates a few times.

{\bf \noindent Backward Operation.}
To use $\text{Prox}_{\text{TV}}^{1D}(\vx, \lambda)$ as a layer, we need to compute the Jacobian with respect to (\wrt) $\vx$ and $\lambda$:
For readability, we let  $\vy=\text{Prox}_{\text{TV}}^{1D}$, which yields
\bea\label{eq:backward}
\frac{\partial \vy}{\partial \vx}  = \mM\mL_{:, \bar\gS}^\intercal ~~\text{and}~~ \frac{\partial \vy}{\partial \lambda} = -\mM\text{Sign}(\mD_N\vy)_{\bar\gS},
\eea
where 
\be\label{eq:grad_m}
\mM=\mL_{:, \bar\gS}(\mL_{:, \bar\gS}^{\intercal}\mL_{:, \bar\gS})^{-1}.
\ee 
Here, $\bar\gS$ denotes the set of indices of non-zero values in $\mD_N\vy$, $\mL$ denotes a $N \times N$ lower triangular matrix, and the subscript denotes the sub-selection of the column. We refer readers to~\citet{cherkaoui2020learning}'s appendix G.1 for the derivation.

To efficiently compute these Jacobian matrices, observe that $\mL_{:, \bar\gS}^{\intercal}\mL_{:, \bar\gS}$ in~\equref{eq:grad_m} is a positive (semi-definite) matrix. Therefore, we use Cholesky factorization and Cholesky solve to compute the inverse instead of a standard matrix inversion. Again, we implemented custom CUDA kernels to allow for efficient indexing and batching. We integrate this into the backward function of PyTorch to support automatic differentiation.

\begin{algorithm}[t]
\caption{$\text{Prox}_{\text{TV}}^{2D}(\mX, \lambda)$ with Proximal Dykstra Method}\label{alg:prox_2d}
\begin{algorithmic}
\STATE {\bf Inputs:} $\mX$ and $\lambda$ \STATE {\bf Initialize:} $\mY^{(k)}= \mX$, $\mP_0=\mathbf{0}$, $\mQ_0=\mathbf{0}$
\FOR{$k \in \{1, \hdots, K\}$}
\FOR[Parallelized with CUDA]{$m \in \{1, \hdots, M\}$}
\STATE $\mZ^{(k)}_{\tt row(m)} = \text{Prox}_{\text{TV}}^{1D}(\mY^{(k)}_{\tt row(m)}+\mP^{(k)}_{\tt row(m)}, \lambda)$ 
\ENDFOR
\STATE $\mP^{(k+1)} = \mP^{(k)} + \mY^{(k)} - \mZ^{(k)}$
\FOR[Parallelized with CUDA]{$n \in \{1, \hdots, N\}$}
\STATE $\mY^{(k+1)}_{\tt col(n)} = \text{Prox}_{\text{TV}}^{1D}(\mZ^{(k)}_{\tt col(m)}+\mQ^{(k)}_{\tt col(n)}, \lambda)$ 
\ENDFOR
\STATE $\mQ^{(k+1)} = \mQ^{(k)} + \mZ^{(k)} - \mY^{(k+1)}$
\ENDFOR
\RETURN $\mY^{K+1}$
\end{algorithmic}
\end{algorithm}

{\bf \noindent 2D TV Proximity.}
The 2D TV proximity implementation is based on the Proximal Dykstra method~\cite{combettes2011proximal}. It alternates between solving 1D TV proximity problems for all the rows and columns, \ie, 
\bea\label{eq:tv1_row}
\min_\mY \frac{1}{2}\norm{\mY-\mX}_F^2 &+&  \lambda \sum_m \norm{\mD_M \mY_{\tt row(m)}}_1\\%
\label{eq:tv1_col}
\min_\mY \frac{1}{2}\norm{\mY-\mX}_F^2 &+& \lambda \sum_n \norm{\mD_N \mY_{\tt col(n)}}_1.
\eea
Both \equref{eq:tv1_row} and \equref{eq:tv1_col} can be decomposed into 1D TV proximity problems per row or per column. As our 1D TV proximity operator implementation supports batching, we can solve all the rows or all the columns in parallel very efficiently on the GPU. The overall procedure is summarized in~\algref{alg:prox_2d}. We found three or four iterations of the Proximal Dykstra method to work well in practice. For the backward pass, our TV 1D proximity operator supports automatic differentiation, so back-propagation through~\algref{alg:prox_2d} is automatically computed with PyTorch.

%% file: figures/fig_pseudo_code.tex
\begin{figure}[t]
{\small
\begin{minted}[frame=lines]{python}
class TVLayer(Module):
  def __init__(self, num_chan, is_sharp):
    # num_chan: Number of channels
    # is_sharp: Sharpen or not.
    self.is_sharp = is_sharp
    self._lmbd = Parameters(zeros(num_chan))
  def forward(self, x):
    # x: Tensor (num_chan, height, width)
    lmbd = softplus(self._lmbd)
    # Apply batched per channel prox. tv
    y = tv_prox_2d(x, lmbd)
    if self.is_sharp:
      y = 2*x-y
    return y
\end{minted}
}
\vspace{-0.3cm}
\caption{Pseudo code of the proposed TV Layer.}
\label{fig:pseudo_code}
\vspace{-0.2cm}
\end{figure}

%% file: sec_exp.tex
\input{results/tab_time_analysis}
\section{Experiments}\label{sec:exp}
First, we compare the running time of TV layers implemented with different approaches. Next, we evaluate the proposed TV layer on a variety of computer vision tasks including: image classification, weakly supervised object localization, edge detection, edge-aware filtering, and image denoising. These tasks cover a wide spectrum of vision applications from high-level semantics understanding to low-level pixel manipulation, demonstrating the practicality of the proposed TV layer. As we are reporting over multiple tasks and metrics, we use $\uparrow$/$\downarrow$ to indicate whether a metric is better when it is higher/lower.

\input{results/tab_cifar10_ood}

\input{subsec_exp/subsec_timing}

\input{subsec_exp/subsec_classification}

\input{subsec_exp/subsec_weakly}

\input{subsec_exp/subsec_edge_detect}

\input{subsec_exp/subsec_edge_preserve}

\input{subsec_exp/subsec_denoise}

%% file: results/tab_time_analysis.tex
\begin{table}[t]
\centering
\setlength{\tabcolsep}{2pt}
\small
\resizebox{0.99\linewidth}{!}{
\begin{tabular}{cc|ccc}
\specialrule{.15em}{.05em}{.05em}
Package & Hardware &Forward & Backward & Total\\
\hline
CVXPYLayers  & CPU & $20704\pm32$ & $9932 \pm 41$ & $1770\times$\\
ProxTV-TS    & CPU & $207.8\pm7.5$ & $430.7\pm7.9$ & $37\times$\\
ProxTV-PN    & CPU & $257.3\pm5.8$ & $447.0\pm9.6$ & $41\times$\\
Ours-PN      & Titan X & $9.0\pm0.7$ & $17.7\pm2.4$ & $1.5\times$\\
Ours-PN      & A6000 & $10.0\pm1.5$ & $7.3 \pm 6.0$ & $1\times$\\
\specialrule{.15em}{.05em}{.05em}
\end{tabular}
}
\vspace{-0.2cm}
\caption{Running time (ms) comparison of TV 1D Proximity Operator. We report the total running time relative to Ours-PN on an A6000 GPU.}
\label{tab:time_analysis}
\vspace{-0.2cm}
\end{table}

%% file: results/tab_cifar10_ood.tex
\begin{table*}[t]
\resizebox{\textwidth}{!}{
\centering
\setlength{\tabcolsep}{2pt}
\begin{tabular}{c|ccc|cccc|cccccccc|c}
\specialrule{.15em}{.05em}{.05em}
& \multicolumn{3}{c|}{Noise Type} & 
\multicolumn{4}{c|}{Blur Type} & 
\multicolumn{8}{c|}{Corruption Type}
\\
Arch. & Gaussian & Shot & Impulse & Glass &Defocus & Motion & Zoom & Snow & Frost & Fog & Brightness & Contrast & Elastic & Pixelate & JPEG & All\\
\hline
AllConv & $36.2$ {\scriptsize $\!\pm\!1.4$} & $48.9$ {\scriptsize $\!\pm\!1.5$} & $\bf 46.6$ {\scriptsize $\!\pm\!3.1$} 
		& $77.9$ {\scriptsize $\!\pm\!0.7$} & $46.7$ {\scriptsize $\!\pm\!2.3$} & $72.8$ {\scriptsize $\!\pm\!0.5$}
		& $71.9$ {\scriptsize $\!\pm\!1.1$} & $77.8$ {\scriptsize $\!\pm\!0.1$} & $71.9$ {\scriptsize $\!\pm\!0.8$} 
		& $82.5$ {\scriptsize $\!\pm\!0.4$} & $91.6$ {\scriptsize $\!\pm\!0.2$} & $67.1$ {\scriptsize $\!\pm\!0.9$}
		& $\bf 82.2$ {\scriptsize $\!\pm\!0.7$} & $72.2$ {\scriptsize $\!\pm\!0.5$} & $\bf 79.0$ {\scriptsize $\!\pm\!0.3$} & $68.4$ {\scriptsize $\!\pm\!0.5$}\\
TV-Smooth & \bf $\bf 42.7 $ {\scriptsize $\!\pm\! 2.4$} & $\bf 55.0 $ {\scriptsize $\!\pm\! 1.4$} & $44.8 $ {\scriptsize $\!\pm\! 3.2$} & $79.4$ {\scriptsize $\!\pm\! 0.7$} & $49.2 $ {\scriptsize $\!\pm\! 2.5$} & $74.5 $ {\scriptsize $\!\pm\! 0.8$} & $\bf 74.3$ {\scriptsize $\!\pm\!1.2$} & $79.4$ {\scriptsize $\!\pm\!0.8$} & $\bf 75.3$ {\scriptsize $\!\pm\!1.1$} & $85.0$ {\scriptsize $\!\pm\! 0.3$} & $92.1 $ {\scriptsize $\!\pm\! 0.1$} & $69.4 $ {\scriptsize $\!\pm\! 1.1$} & $81.8 $ {\scriptsize $\!\pm\! 0.3$} & $\bf 73.0$ {\scriptsize $\!\pm\!0.6$} & $78.4$ {\scriptsize $\!\pm\!0.2$} & $70.3$ {\scriptsize $\!\pm\!0.6$}\\
TV-Sharp & $40.6$ {\scriptsize $\!\pm\!1.4$} & $53.1$ {\scriptsize $\!\pm\!1.5$} & $46.5$ {\scriptsize $\!\pm\!1.0$} & $\bf 79.8$ {\scriptsize $\!\pm\!0.3$} & $\bf 50.3$ {\scriptsize $\!\pm\!0.6$} & $\bf 75.7$ {\scriptsize $\!\pm\!0.7$} & $73.9$ {\scriptsize $\!\pm\!0.4$} & $\bf 80.3$ {\scriptsize $\!\pm\!0.5$} & $75.2$ {\scriptsize $\!\pm\!0.9$} & $\bf 86.2$ {\scriptsize $\!\pm\!0.4$} & $\bf 92.2$ {\scriptsize $\!\pm\!0.1$} & $\bf 74.4$ {\scriptsize $\!\pm\!0.2$} & $81.8$ {\scriptsize $\!\pm\!0.3$} & $72.6$ {\scriptsize $\!\pm\!0.5$} & $76.5$ {\scriptsize $\!\pm\!0.16$} & $\bf 70.6$ {\scriptsize $\!\pm\!0.2$}\\
\specialrule{.15em}{.05em}{.05em}
\end{tabular}
}%
\vspace{-0.2cm}
\caption{Classification accuracy ($\uparrow$) on CIFAR10-C over different types of corruptions.
}
\label{tab:cifar10_ood_result}
\vspace{-0.2cm}
\end{table*}

%% file: subsec_exp/subsec_timing.tex
\subsection{Timing Analysis}\label{subsec:timing}
We compare with a generic TV solver using
CVXPYLayers~\cite{Diamond_Boyd_2016, Agrawal_Amos_Barratt_Boyd_Diamond_Kolter_2019} which supports back-propagation. We also compare with specialized TV solvers using the ProxTV toolbox\footnote{Available at \url{https://github.com/albarji/proxTV}.}~\cite{jimenez2011fast,Barbero_Sra_2018} with a PyTorch implementation of the backward pass. Both, CVXPYLayers and ProxTV only support CPU computations.

We time each of these methods on a batch of signals with dimension $256 \times 32 \times 32$. This is a typical dimension for small scale computer vision tasks. The data contains a unit step signal with additive Gaussian noise and $\lambda$ is set to one. Timing evaluation  is done on an NVIDIA TITAN X or A6000 GPU and an Intel Core i7-6700K CPU. We report the mean and standard deviation over 25 runs.

{\noindent\bf Results.} In~\tabref{tab:time_analysis} we report the forward and backward computation time, in milliseconds, for each of the methods. We observe that ProxTV with specialized solver is faster than CVXPYLayers using a generic solver. We report ProxTV with two different specialized TV solvers, namely, Taut-String (TS) and Projected Newton (PN). However, these CPU based methods remain too slow for practical vision applications, \eg, an entire inference on ResNet101 takes $\sim150$ ms. With GPU support, our approach achieves a $1770\times$ speedup over CVXPYLayers, and is $37\times$ to $41\times$ faster than ProxTV. This significant speedup enables the scaling of TV layers to real computer vision tasks.

%% file: subsec_exp/subsec_classification.tex
\input{figures/fig_class_lmbd_vs_iter}
\subsection{Image Classification}\label{subsec:classification}
We conduct experiments on CIFAR10 data~\cite{Krizhevsky_Hinton_others_2009} evaluating two aspects: (a) standard classification and (b) out of domain generalization.
We study the effect of the proposed total variation layer on a baseline architecture of an All Convolutional Network~\cite{springenberg2014striving}. We study both the smoothing and the sharpening TV layer, with $\lambda$ shared across channels, being added in the first three convolution blocks. Specifically, we insert the TV layers before batch-normalization of each block. We refer to these modified architectures as TV-Smooth and TV-Sharp. We initialize $\lambda=0.05$ and train them jointly with the model parameters. 

{\noindent\bf Standard Classification.}
On CIFAR10, the baseline model AllConvNet achieved an accuracy of $93.51\%\pm 0.18\%$ (standard deviation reported over five runs). With the smoothing and sharpening TV layer added, TV-Smooth and TV-Sharp achieve an accuracy of $93.61\pm0.17\%$ and $93.43\pm0.24\%$ respectively. As can be seen, adding these TV layers performs on par with the baseline model.

To ensure that the models actually use the TV modules, we visualize $\lambda$ in~\equref{eq:prox_tv1d}, at each layer, throughout training iterations in~\figref{fig:class_lmbd_vs_iter}. Note, $\lambda$ learns to be non-zero, indicating that the TV layers are impacting the model architecture.

{\noindent\bf Out of Domain Generalization.} To further assess how TV modules affect the model, we evaluate out-of-distribution (o.o.d.) generalization using CIFAR10-C data~\cite{hendrycks2019benchmarking}. Models are trained on clean data and evaluated on corrupted data, \eg, noise is added to the images. Classification accuracy on CIFAR10-C are reported in~\tabref{tab:cifar10_ood_result}. On average, both the TV-Smooth and TV-Sharp model improve the o.o.d.\ generalization over the baseline model. We observe: the TV-Smooth model generalizes better to additive noise corruption, whereas TV-Sharp improves the accuracy on blur corruption. These results match our expected behavior of sharpening and smoothing and illustrate that optimization as a layer is a useful way to encode a models' inductive bias.

%% file: figures/fig_class_lmbd_vs_iter.tex
\begin{figure}[t]
\vspace{-0.05cm}
\centering
\includegraphics[width=0.49\linewidth]{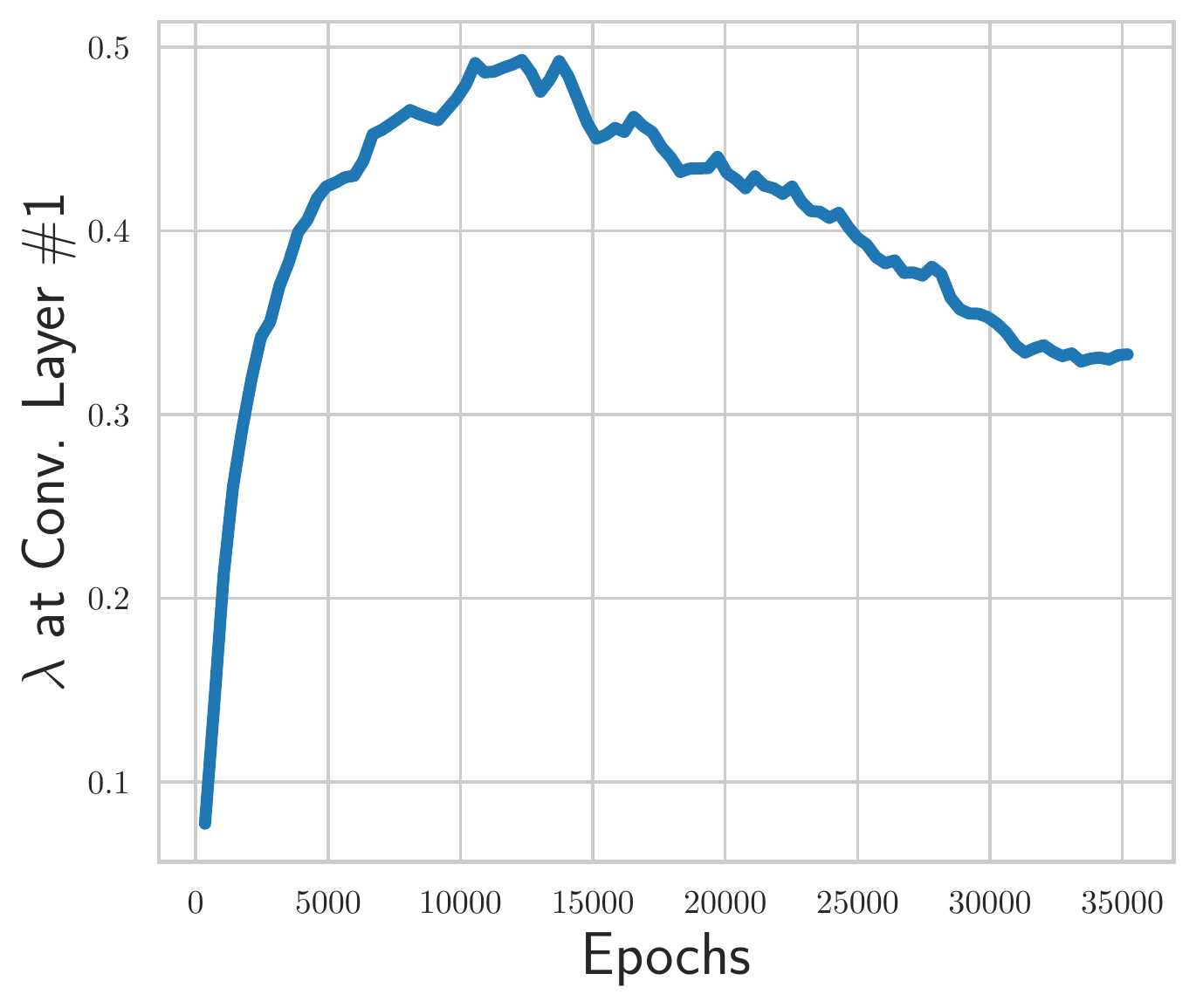}
\includegraphics[width=0.49\linewidth]{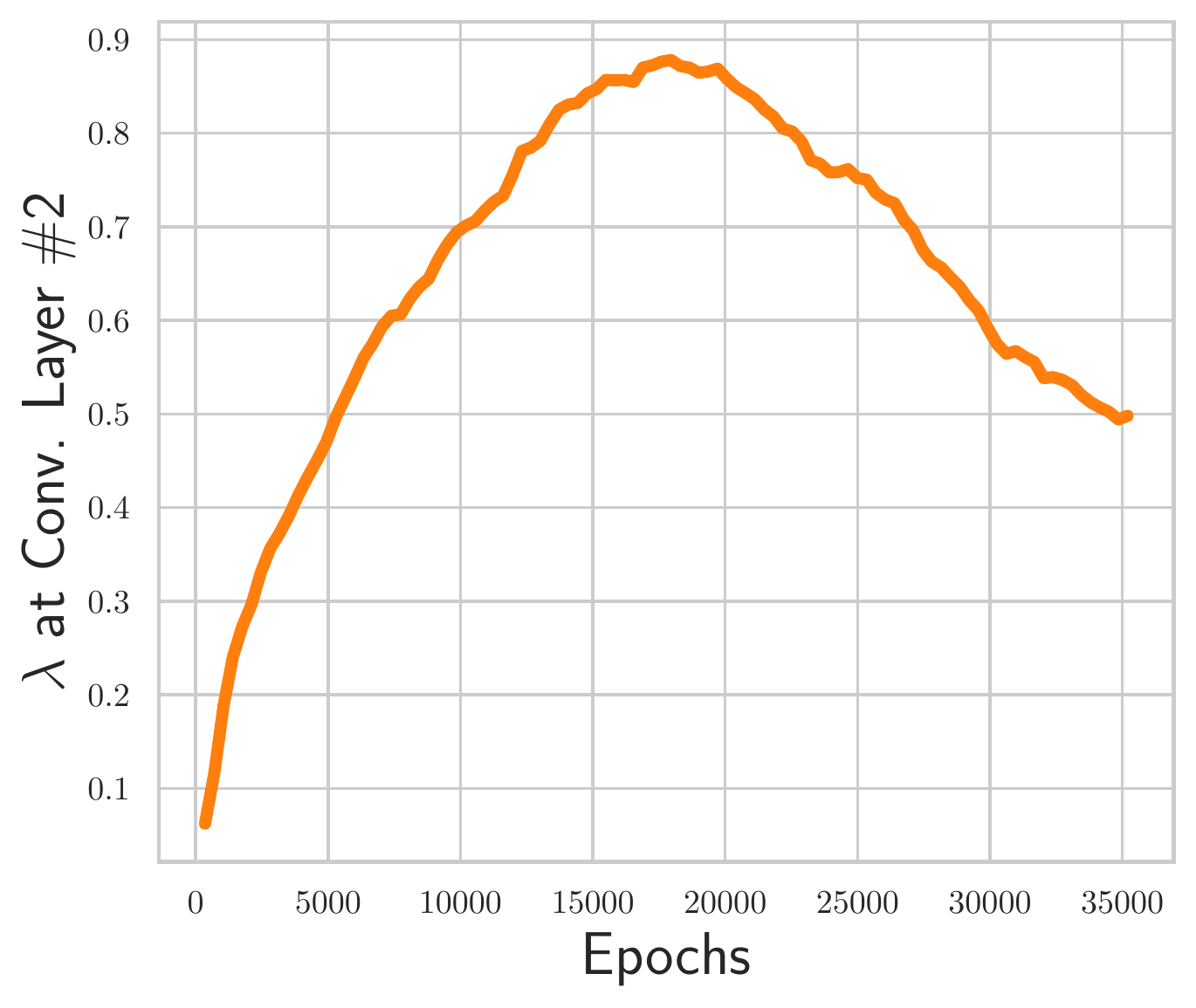}
\vspace{-0.2cm}
\caption{Visualization of $\lambda $ \vs number of training epochs for TV-Sharp at the first and second Conv.\ block. We observe that $\lambda$ learns to be greater than zero.}
\label{fig:class_lmbd_vs_iter}
\vspace{-0.2cm}
\end{figure}

%% file: subsec_exp/subsec_weakly.tex
\begin{table}[t]
\centering
\resizebox{0.99\linewidth}{!}{
\setlength{\tabcolsep}{3pt}
\begin{tabular}{c|ccc}
\specialrule{.15em}{.05em}{.05em}
Method & VGG-16 & Inception-V3 & ResNet-50 \\
\hline
CAM-paper & 60.02 & 63.40 & 63.65 \\
CAM-repro. & 60.13 & 63.51 & 64.09 \\
CAM-repro.+TV (Ours) & \textbf{60.35} & \textbf{63.80} & \textbf{65.36} \\
\specialrule{.15em}{.05em}{.05em}
\end{tabular}}
\vspace{-0.1cm}
\caption{WSOL localization accuracy, \texttt{MaxBoxAccV2} ($\uparrow$), on the test set from~\citet{choe2020cvpr} for ImageNet pre-trained models.}
\label{tab:wsol}
\vspace{-0.1cm}
\end{table}

\begin{figure}[t]
\centering
\includegraphics[width=\linewidth]{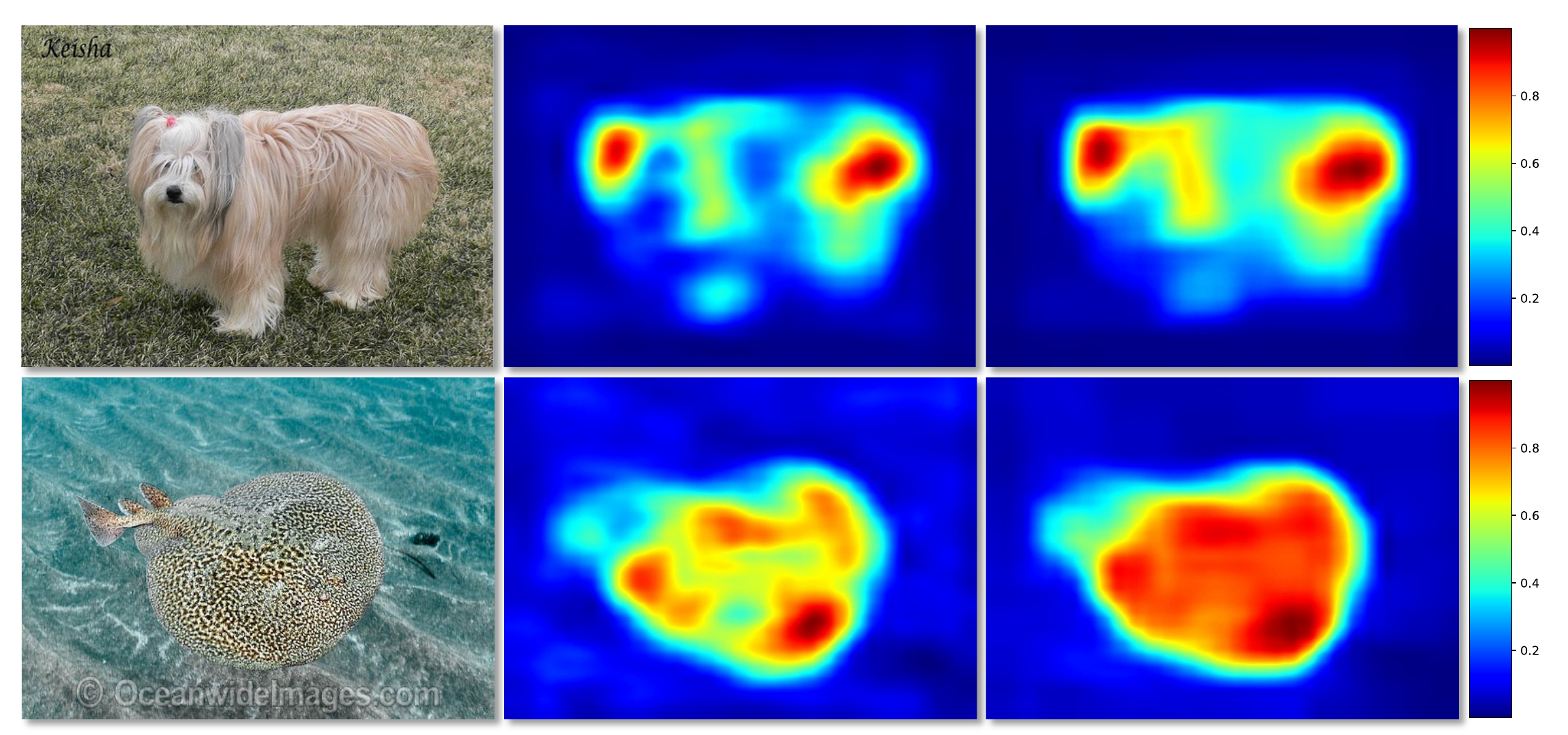}
\vspace{-0.5cm}
\caption{Visualization of WSOL results. We show the input images (left),  the heat-maps from CAM (center) and ours (right). }
\label{fig:wsol}
\vspace{-0.2cm}
\end{figure}

\subsection{Weakly Supervised Object Localization}\label{subsec:weakly}
Weakly Supervised Object Localization (WSOL) is a popular interpretability tool in various computer vision tasks. It learns to localize objects with only image-level labels. The seminal work Class Activation Mapping (CAM)~\cite{zhou2016learning} first studies WSOL for image classification. Follow-up work~\cite{selvaraju2017grad, ren2021wypr} further generalizes to broader domains such as vision and language. We believe TV layer is beneficial to WSOL as it aids the localization results, \ie, class-wise heat-maps, to be smoother and better aligned with  boundaries.

For evaluation, the most popular method is to infer surrounding bounding boxes of the computed class heat-maps and compare to ground-truth ones. However, recent WSOL analysis work~\cite{choe2020cvpr} points out that CAM is still the state-of-the-art WSOL method under a fairer evaluation setting. The performance boost achieved in follow-ups  are illusory caused by wrong experimental settings and inconsistent bounding box generation methods. We thus adopt both CAM and the fair evaluation protocol in this work and test TV layer on it.

{\bf\noindent Experimental Setup.}
We test three different models: VGG-16~\cite{simonyan2014very}, ResNet-50~\cite{he2016deep}, and Inception-V3~\cite{szegedy2016rethinking}. 
We use a fixed TV2D-Smooth layer shared across all channels with $\lambda=1$. We insert this TV layer right before the CAM layer of each network.
We use the code base from~\citet{choe2020cvpr} who discuss a more fair experimental setting: all WSOL methods are fine-tuned on a fixed validation set to search for the best hyper-parameters and then test on newly collected test images. Since the fine-tuned models are not released, we reproduce using the released code and report both results (CAM-repro.\ and CAM-paper) for completeness. 

{\bf\noindent Results.}  
We report quantitative metrics in~\tabref{tab:wsol} where we observe that a TV layer consistently improves WSOL results of various models (0.22/0.29/1.27 for VGG-16/Inception-V2/ResNet-50).
We further show qualitative comparisons between the vanilla version and ours in~\figref{fig:wsol} where we observe that the TV layer helps WSOL models to generate smoother and aligned results.

%% file: subsec_exp/subsec_edge_detect.tex
\input{results/tab_edge_results}

\subsection{Edge Detection}\label{subsec:edge_detect}
Edge detection is the task of identifying all the edges for a given input image. For learning based methods, this is formulated as a binary classification problem for each pixel location, \ie, classifying whether a given pixel in the image is an edge. The task is illustrated  in~\figref{fig:edge_detect_qual}.
We evaluate on the recent BIPED~\cite{Poma_Riba_Sappa_2020} and 
the Multicue (MBDB)~\cite{Mely_Kim_McGill_Guo_Serre_2016} dataset for edge detection.

{\noindent \bf Experimental Setup.} We compare to the  Dense extreme inception Network for edge detection (DexiNed) baseline proposed by~\citet{Poma_Riba_Sappa_2020}. DexiNed consists of convolutional blocks and follows a multi-scale and multi-head architecture. Before each stage of max-pooling, DexiNed outputs an edge-map at the scale of $2\times$, $4\times$, $8\times$ and $16\times$. The final edge-map prediction is obtained by averaging the edge-map at each scale. For our model, we added TV2D-Sharp layers, with trainable $\bm\lambda$, at the $2\times$, $4\times$ and $8\times$ edge-maps. At the final edge-map, we added a TV2D-Smooth layer.

Evaluation metrics for edge detection~\cite{arbelaez2010contour} are based on the 
\be
\text{F-measure} = 2 \cdot \frac{\text{precision} \cdot \text{recall}}{\text{precision} + \text{recall}},
\ee
with different thresholds, including: (a) optimal dataset scale (ODS) which corresponds to using the best threshold over a dataset; (b) optimal image scale (OIS) which corresponds to using the best threshold per image and average precision (AP) which is the area under the precision-recall curve.

{\bf\noindent Results.} 
Following~\citet{Poma_Riba_Sappa_2020}, we compared our DexiNed + TV to the standard DexiNed~\cite{Poma_Riba_Sappa_2020}, RCF~\cite{Liu_Cheng_Hu_Wang_Bai_2017} and HED~\cite{Xie_Tu_2015} in~\tabref{tab:edge_results} using the BIPED and MDBD dataset. All the models are trained only on the BIPED dataset. 
As can be seen, our approach achieves improvements across ODS, OIS, and AP for both BIPED and MDBD data. A qualitative comparison of the final detected edges is shown in~\figref{fig:edge_detect_qual}. We observe that DexiNed+TV predicts sharper edges and suppresses textures better than DexiNed.

\input{figures/fig_edge_detect_qual}

%% file: results/tab_edge_results.tex
\begin{table}
\centering
\setlength{\tabcolsep}{2pt}
\resizebox{0.99\linewidth}{!}{
\begin{tabular}{c|ccc|ccc}
\specialrule{.15em}{.05em}{.05em}
&  \multicolumn{3}{c|}{BIPED~\cite{Poma_Riba_Sappa_2020}} 
&  \multicolumn{3}{c}{MDBD~\cite{Mely_Kim_McGill_Guo_Serre_2016}}\\
Method & ODS ($\uparrow$) & OIS ($\uparrow$) & AP ($\uparrow$) & ODS ($\uparrow$) & OIS ($\uparrow$) & AP($\uparrow$)\\
\hline
HED~\cite{Xie_Tu_2015} & .829 & .847 & .869 & .851 & .864 & .890\\
RCF~\cite{Liu_Cheng_Hu_Wang_Bai_2017} & .843  & .859 & .882 & .857 & .862 & -\\
DexiNed~\cite{Poma_Riba_Sappa_2020} & .859  & .867 & .905 & .859  & .864 & .917\\
DexiNed+TV &  \bf .874 & \bf .879 & \bf .914 & \bf .863 & \bf .875 & \bf .920\\
\specialrule{.15em}{.05em}{.05em}
\end{tabular}
}
\vspace{-0.2cm}
\caption{Quantitative comparison for edge detection on BIPED and MDBD data. Baselines are obtained from corresponding papers.}
\label{tab:edge_results}
\vspace{-0.2cm}
\end{table}

%% file: figures/fig_edge_detect_qual.tex
\begin{figure}[t]
\centering
\small
\setlength{\tabcolsep}{1pt}
\renewcommand{\arraystretch}{0.8}
\begin{tabular}{cc}
\includegraphics[width=0.49\linewidth]{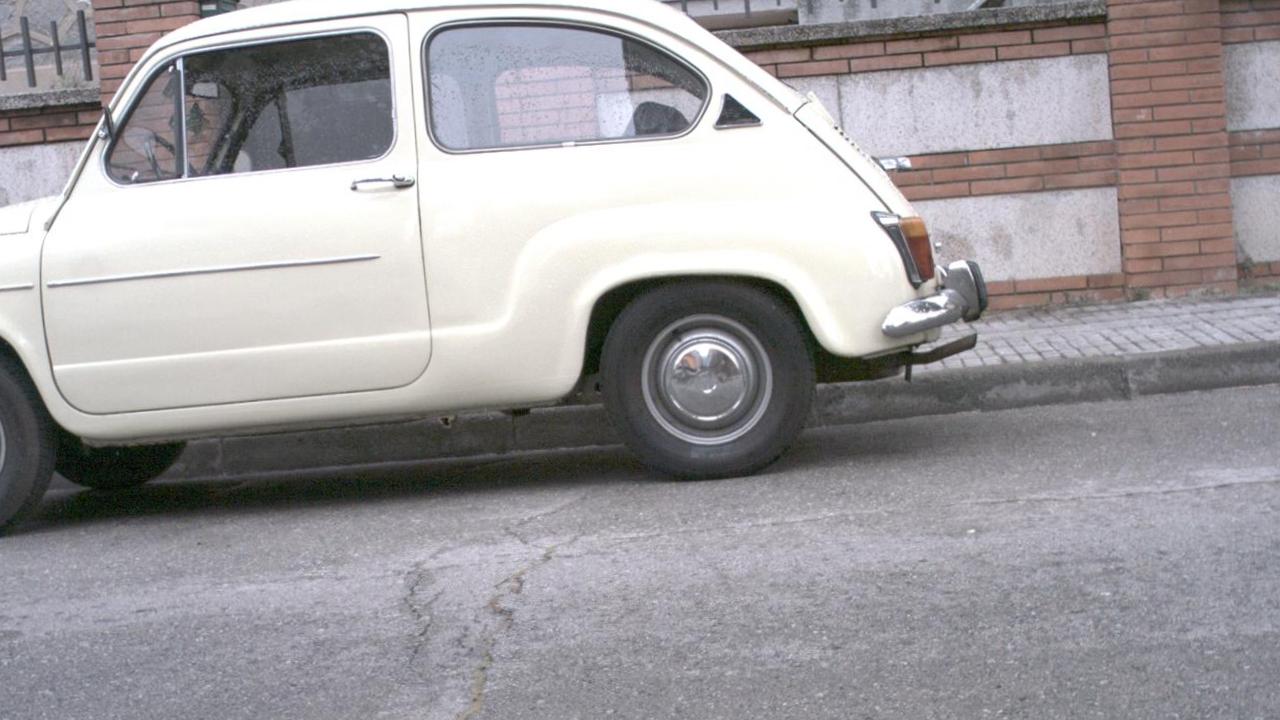} &
\includegraphics[width=0.49\linewidth]{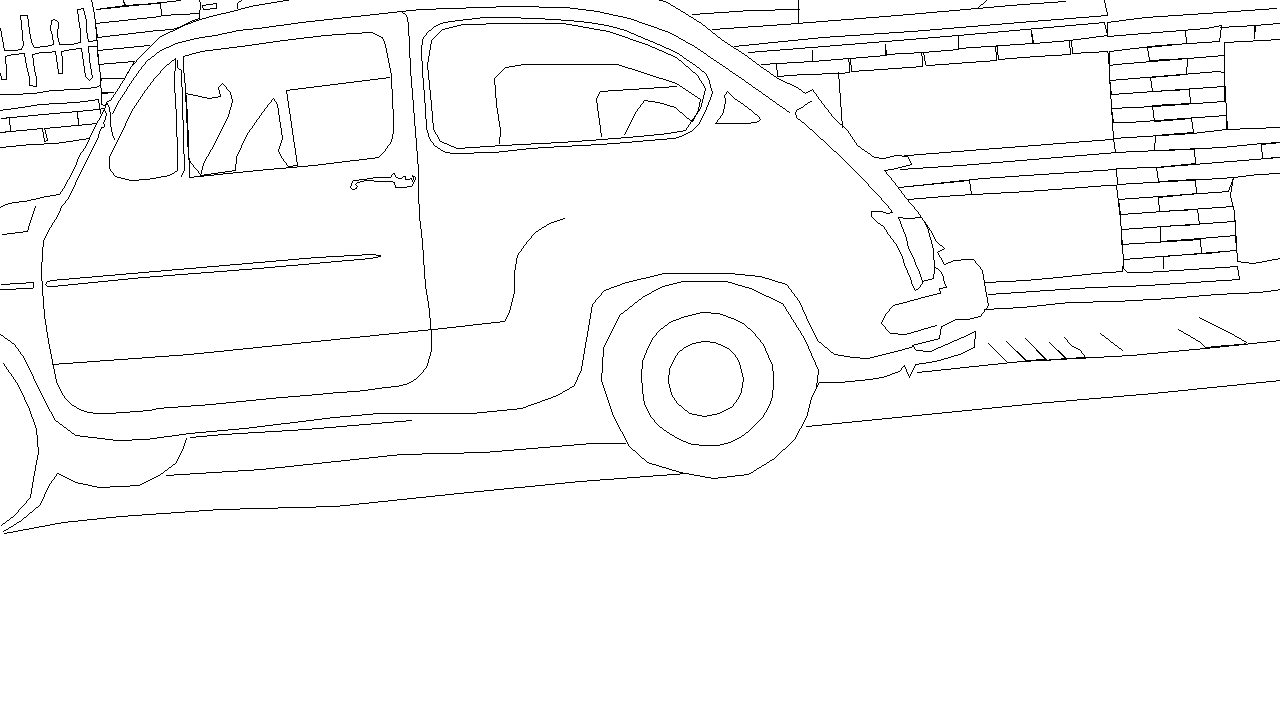}\\
Input & Ground-Truth\\
\includegraphics[width=0.49\linewidth]{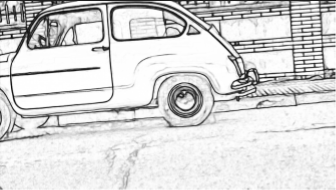} &
\includegraphics[width=0.49\linewidth]{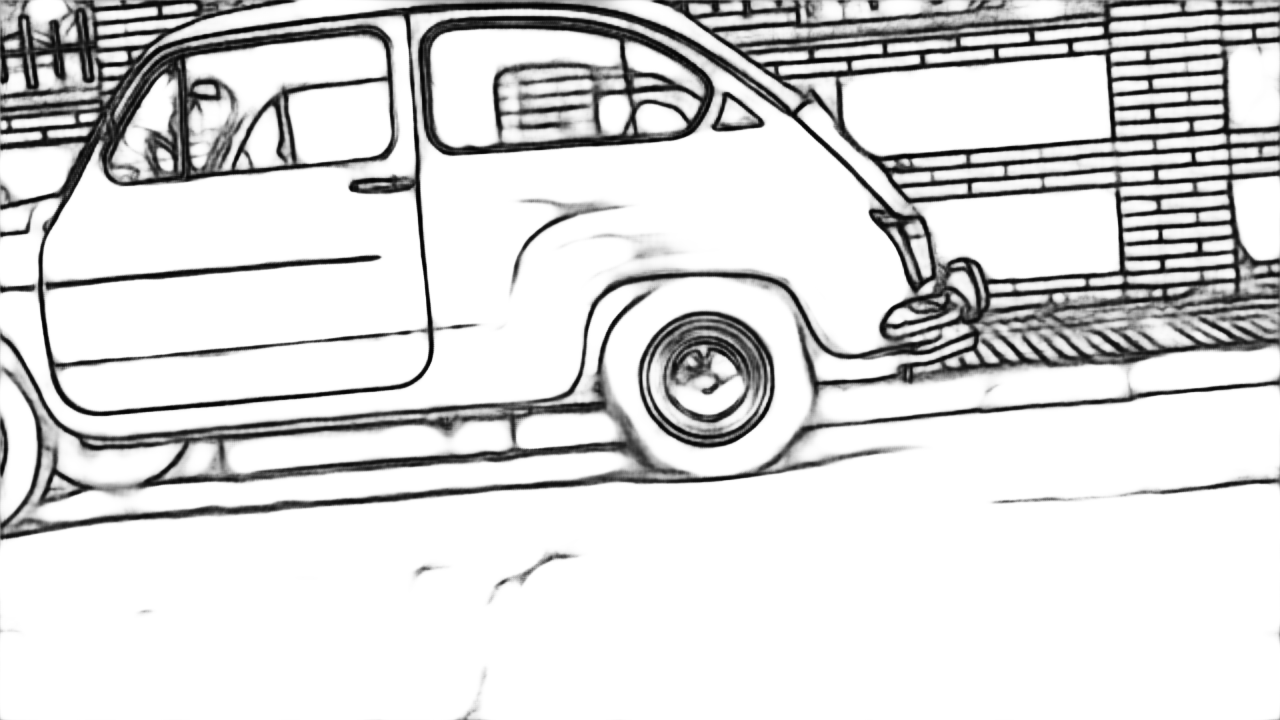}\\
DexiNed~\cite{Poma_Riba_Sappa_2020} & DexiNed+TV (Ours)
\end{tabular}
\vspace{-0.15cm}
\caption{Qualitative comparison on edge detection.}
\vspace{-0.2cm}
\label{fig:edge_detect_qual}
\end{figure}

%% file: subsec_exp/subsec_edge_preserve.tex
\begin{figure*}[t]
    \small
    \centering
    \begin{tabular}{cccc}
    \includegraphics[width=0.285\linewidth]{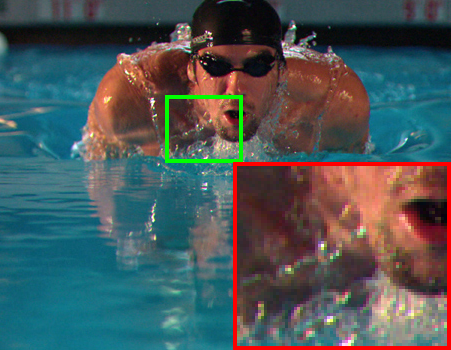} &
    \includegraphics[width=0.285\linewidth]{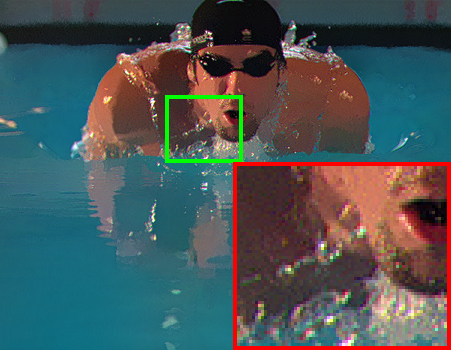} &
    \includegraphics[width=0.285\linewidth]{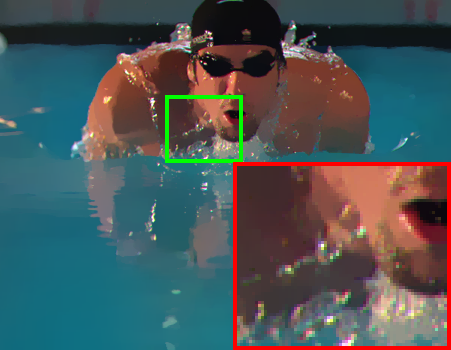}\\
    \includegraphics[width=0.285\linewidth, 
                    trim={0 2cm 0 0},clip]{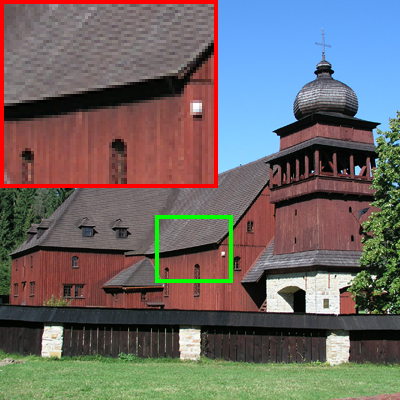} &
    \includegraphics[width=0.285\linewidth,
                    trim={0 2cm 0 0},clip]{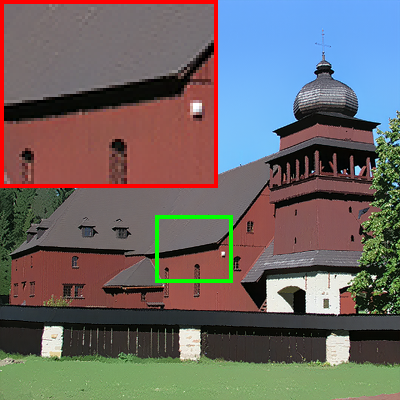} &
    \includegraphics[width=0.285\linewidth,
                    trim={0 2cm 0 0},clip]{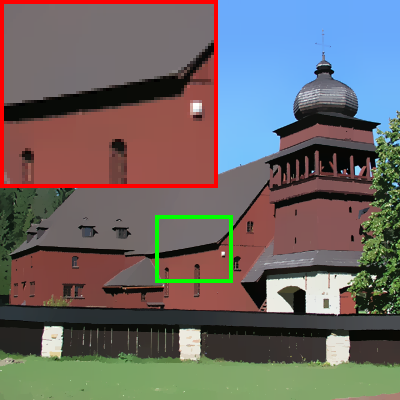}\\
    Input & Final Conv.\ Layer & After TV-Smooth
    \end{tabular}
    \vspace{-0.2cm}
    \caption{Visualization of the effects of the final TV-smooth layer for edge preserving smoothing. }
    \label{fig:edge_preserve_out}
    \vspace{-0.3cm}
\end{figure*}

\subsection{Edge Preserving Smoothing}\label{subsec:edge_preserve}
Edge preserving smoothing is the task of smoothing images while maintaining sharp edges. To fairly compare across algorithms, ~\citet{Zhu_Liang_Jia_Zhang_Yu_2019} propose a dataset (BenchmarkEPS) consisting of 500 training and testing images with corresponding ``ground-truth'' smoothed images.  

\citet{Zhu_Liang_Jia_Zhang_Yu_2019} also propose to use two evaluation metrics, Weighted Root Mean Squared Error (WRMSE) and Weighted Mean Absolute Error
(WMAE) defined as follows:
\be\nonumber
\small
\text{WRMSE}(\mI, \hat\mI) = \sqrt{\frac{\sum_{n=1}^N\sum_{k=1}^K w^{(n,k)} \norm{\hat\mI^{(n)} - \mI^{(n,k)}}_2^2
}{N \cdot K \cdot H \cdot W}
}
\ee
\text{~and~} 
\be\nonumber
\small
\text{WMAE}(\mI, \hat\mI) = \frac{\sum_{n=1}^N\sum_{k=1}^K w^{(n,k)} \norm{\hat\mI^{(n)} - \mI^{(n,k)}}_1
}{N \cdot K \cdot H \cdot W},
\ee
where $\mI^{(n,k)}$ corresponds to the $k$-th ground-truth of the $n$-th image (with height $H$ and width $W$), $\hat\mI^{(n)}$ denotes the prediction of the $n$-th image, $w^{(n,k)}$ denotes the normalized weight of the ground-truth. Note, there are multiple ground-truths for each image, hence annotators vote for each ground-truth and their votes are normalized into weight $w^{(n,k)}$. 

 For evaluation, we compared with two deep-net baselines, VDCNN and ResNet, proposed by~\citet{Zhu_Liang_Jia_Zhang_Yu_2019}, as well as an optimization based approach of L1 smoothing~\cite{Bi_Han_Yu_2015}. The ResNet architecture consists of 16 Residual Blocks followed by three convolutional layers with a skip connection from the input. For our model (ResNetTV), we insert four TV1D-Sharp layers with alternating row/column directions into the Residual Blocks and lastly a TV2D-Smooth layer (with shared $\lambda$ across channels) after the last skip connection. This design is due to the observations that residual blocks learn high-frequency content and the final output is smooth.

\input{results/tab_edge_preserve}

\begin{figure}[t]
\vspace{-0.2cm}
    \centering
    \small
    \setlength{\tabcolsep}{0pt}
    \renewcommand{\arraystretch}{0.2}
    \begin{tabular}{cccc}
    \includegraphics[height=2.6cm,
                    trim={0 0 2.4cm 0},clip]{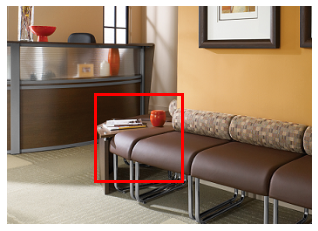} &
    \includegraphics[height=2.6cm, trim={0 0 0.21cm 0},clip]{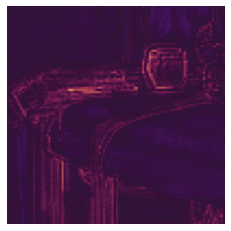} &
    \includegraphics[height=2.6cm]{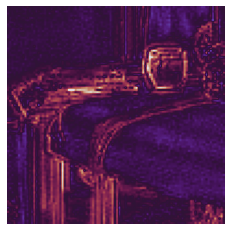} &
    \includegraphics[height=2.6cm]{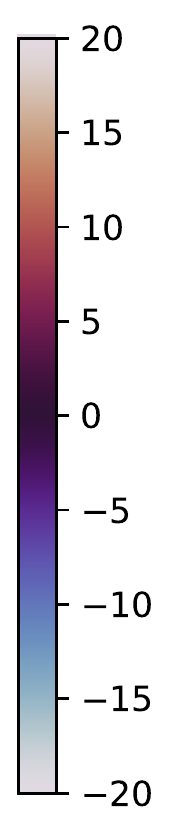}\\
    \includegraphics[height=2.6cm,
                     trim={0 0 2.4cm 0},clip]{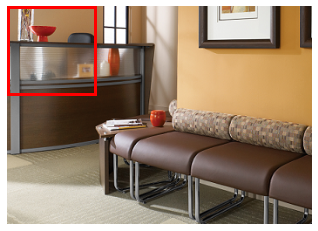} &
    \includegraphics[height=2.6cm, trim={0 0 0.21cm 0},clip]{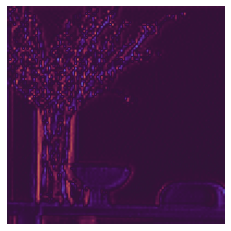} &
    \includegraphics[height=2.6cm]{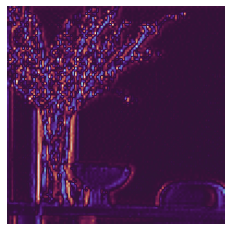} &
    \includegraphics[height=2.6cm]{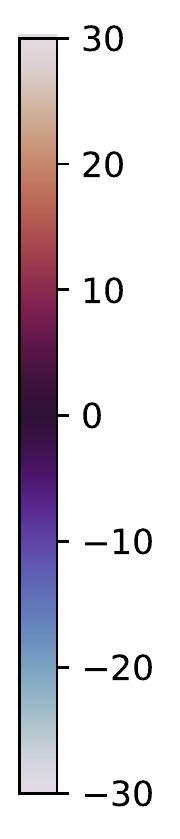}\\
    Input & Feature & After TV-Sharp
    \end{tabular}
    \vspace{-0.2cm}
    \caption{Visualization of feature maps before and after TV-Sharp layer at a residual block on ResNetTV. The red box indicates the region of visualized (zoomed-in) feature maps.}
    \label{fig:edge_preserve_feat}
    \vspace{-0.4cm}
\end{figure}

{\bf\noindent Results.}
In~\tabref{tab:edge_preserve}, we report the quantitative results. We observe improvements in both WRMSE and WMAE over the baselines, both the pure optimization based method and deep learning methods. This demonstrates the benefits of incorporating optimization as a layer into deep-nets.
Beyond the quantitative improvements, we analyze the effect of the final TV-Smooth layer, which can be easily visualized as it is operating in the image space. First, we observe that the learned $\lambda=15.1$, which means that the layer is indeed performing smoothing. To illustrate its effect, in~\figref{fig:edge_preserve_out} we visualize the image at the final convolutional layer and after the TV-Smooth layer. As can be seen, the image at the final convolutional layer (column 2 of~\figref{fig:edge_preserve_out}) is already smoothed by a decent amount. The final TV-Smooth layer further filters the image while preserving the edges which improves the result. We suspect that  addition of the TV-Smooth layer aids the overall performance as the deep-net does not need to use its capacity for part of the smoothing procedure.

\input{figures/fig_denoising_qual}

We also analyze the effect of the TV layer on the feature maps. In~\figref{fig:edge_preserve_feat} we show an intermediate feature map before and after a TV-Sharp layer. As expected,  feature maps are sharpened, leading to more prominent edges in  feature space. While it is difficult to directly interpret these features maps, intuitively, a deep-net that performs well on edge preserving smoothing should  easily capture edges of an image.

%% file: results/tab_edge_preserve.tex
\begin{table}
\centering
\begin{tabular}{c|ccc}
\specialrule{.15em}{.05em}{.05em}
Method & WRMSE ($\downarrow$) & WMAE ($\downarrow$)\\
\hline
L1 smooth~\cite{Bi_Han_Yu_2015} 
& 9.89 & 5.76\\
VDCNN~\cite{Zhu_Liang_Jia_Zhang_Yu_2019} 
&9.78 & 6.15  \\
ResNet~\cite{Zhu_Liang_Jia_Zhang_Yu_2019} 
& 9.03 & 5.55\\
ResNetTV (Ours) &  \bf 8.87 & \bf 5.47\\
\specialrule{.15em}{.05em}{.05em}
\end{tabular}
\vspace{-0.2cm}
\caption{Quantitative results on the edge-preserving image smoothing benchmark~\cite{Zhu_Liang_Jia_Zhang_Yu_2019}.
}
\label{tab:edge_preserve}
\vspace{-0.3cm}
\end{table}

%% file: figures/fig_denoising_qual.tex
\begin{figure*}[t]
    \centering
    \small
    \setlength{\tabcolsep}{1.5pt}
    \begin{tabular}{ccccc}
    \includegraphics[width=0.19\linewidth]{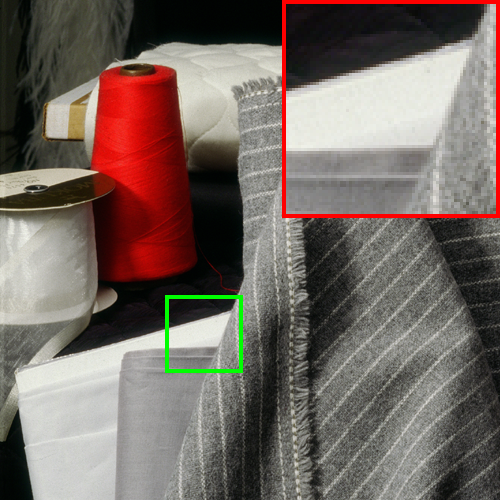} & 
    \includegraphics[width=0.19\linewidth]{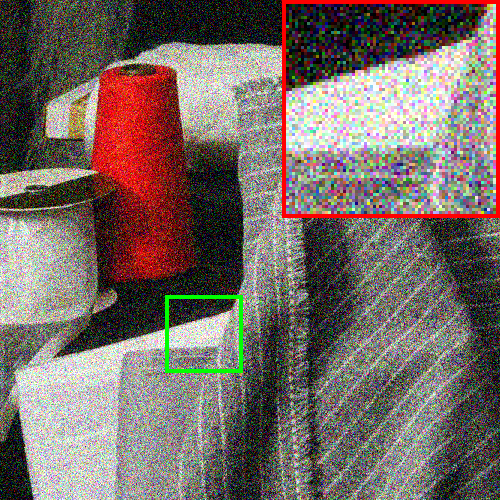} &
    \includegraphics[width=0.19\linewidth]{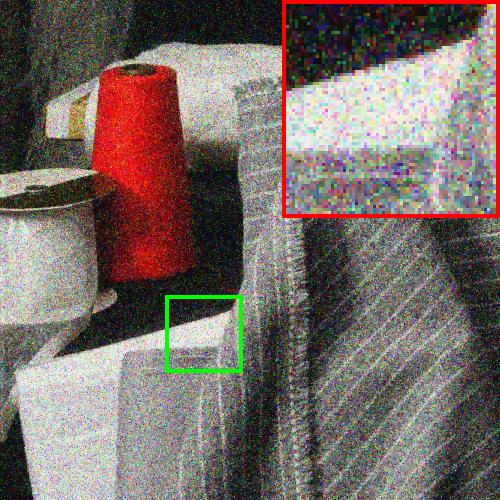} &
    \includegraphics[width=0.19\linewidth]{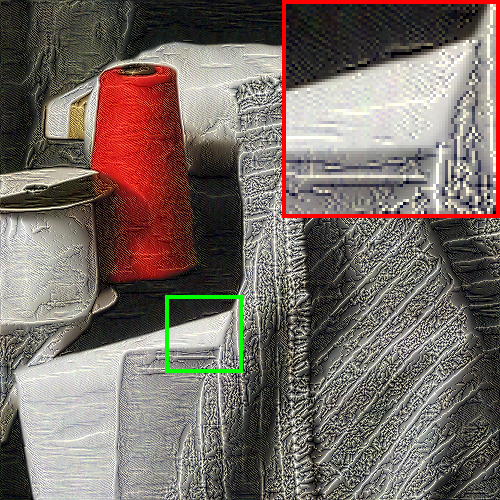} &
    \includegraphics[width=0.19\linewidth]{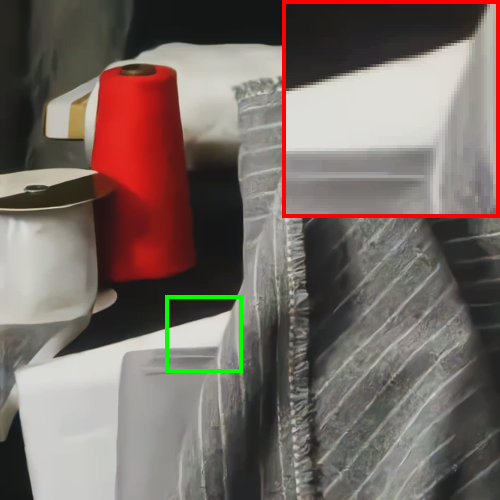}\\
    \includegraphics[width=0.19\linewidth]{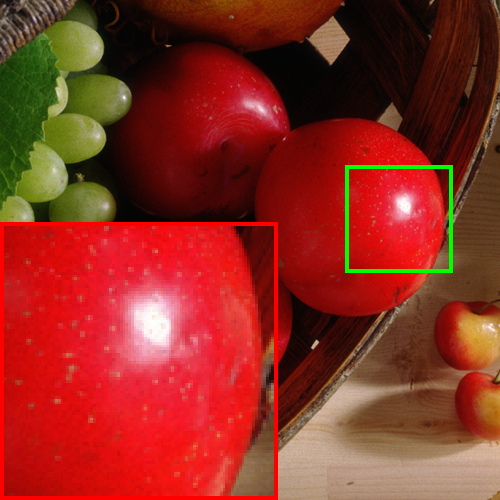} & 
    \includegraphics[width=0.19\linewidth]{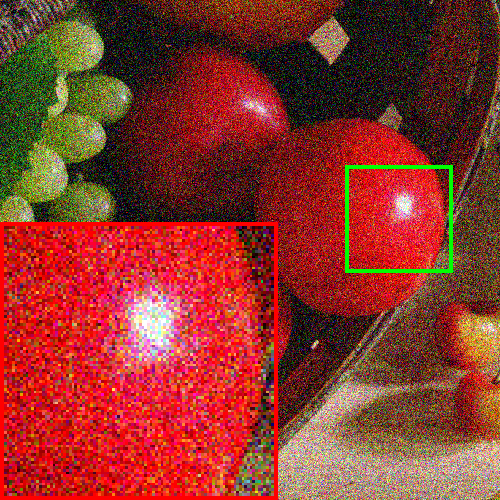} &
    \includegraphics[width=0.19\linewidth]{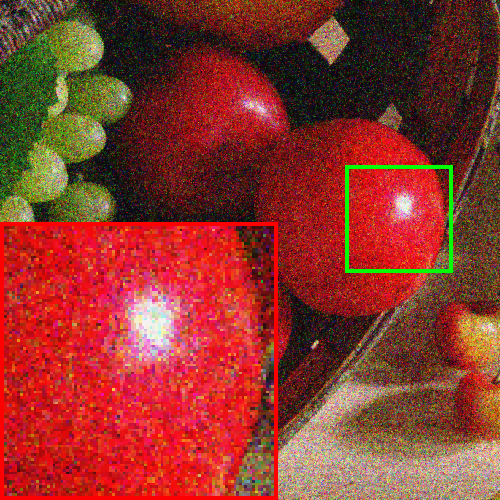} &
    \includegraphics[width=0.19\linewidth]{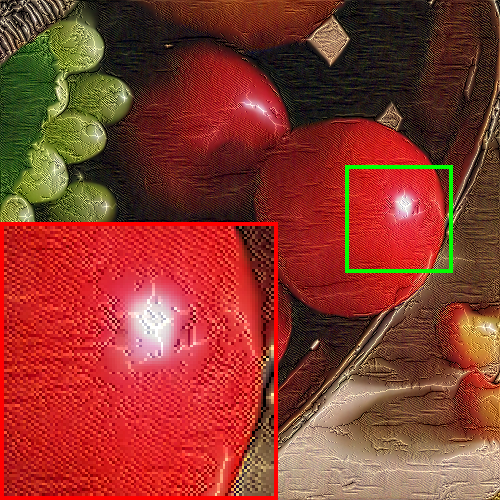} &
    \includegraphics[width=0.19\linewidth]{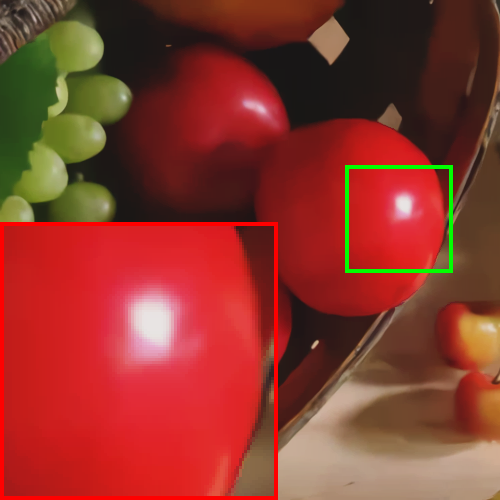}\\
    Ground-Truth & Noisy Input & After Input TV & Final Residual Block & After Output TV
    \end{tabular}
    \vspace{-0.1cm}
    \caption{Visualizing the effects of the input and output TV layers (see~\equref{eq:dncnntv}) on image denoising.}
    \label{fig:qual_denoising}
    \vspace{-0.15cm}
\end{figure*}

%% file: subsec_exp/subsec_denoise.tex
\subsection{Image Denoising}\label{subsec:denoise}
Image denoising is the task of recovering a clean image given a noisy input image. Deep-net-based approaches typically formulate image denoising as a regression task, \ie, regress to the clean RGB values given a noisy input. We consider color images corrupted by additive white Gaussian noise and report the average peak signal-to-noise ratio (PSNR)~\cite{hore2010image}; the higher the better. We evaluate on the common  CBSD68~\cite{Martin_Fowlkes_Tal_Malik_2001}, Kodak24~\cite{Franzen_1999} and McMaster~\cite{Zhang_Wu_Buades_Li_2011} data.

We use DnCNN~\cite{Zhang_Zuo_Chen_Meng_Zhang_2017} as our base model. It consists of residual blocks with a full skip connect from the input, \ie, 
\be
\text{DnCNN}(\mI) \triangleq \gR(\mI)+\mI,
\ee
where $\gR$ denotes the residual blocks and $\mI$ is the noisy input image.
We added TV2D-Smooth layers at the input and output of the network, \ie,
\be\label{eq:dncnntv}
\text{DnCNNTV}(\mI) \triangleq \text{Prox}_{\text{TV}}^{2D}\left(
\left[\gR(\tilde\mI)+ \tilde\mI\right]
, \bm\lambda_{\tt out}\right),
\ee
where $\tilde\mI \triangleq \text{Prox}_{\text{TV}}^{2D}(\mI, \bm\lambda_{\tt in})$.
We use the KAIR toolbox\footnote{Available at \url{https://github.com/cszn/KAIR}} to train these denoising models.

{\bf\noindent Results.}
In~\tabref{tab:denoise_results}, we report the average PSNR over different noise-levels.
We observe a larger gain on the $\sigma=50$ setting and on the McMaster dataset. We further analyze the behavior of the two added TV layers. As the TV layers are added on the image space, we can directly visualize them. In~\figref{fig:qual_denoising} we visualize the noisy input image, the image immediately after the input TV layer, the final residual block, and the final result after the output TV layer. We observe: the input TV layer performs a very weak noise reduction, see column two \vs three in~\figref{fig:qual_denoising}. Next, in column four, we observe: the residual block outputs an image with sharp edges but with high-frequency artifacts. These artifacts are then smoothed by the final output TV layer (see col.\ five).

{\bf \noindent Limitations.} We note that our result is not the state-of-the-art model, \ie, SwinTransformer~\cite{liang2021swinir}. We have also added TV layers to SwinTransformer models. However, $\bm\lambda$s learn to be zero which effectively turns the TV layer off. We suspect that the TV smoothing layer leads to overly smooth  output. Hence, when a deep-net has enough capacity %
the TV layer may learn to use $\bm\lambda=0$ to avoid  smoothing. %

\input{results/tab_denoise_results}

%% file: results/tab_denoise_results.tex
\begin{table}[t]
\small
\centering
\setlength{\tabcolsep}{3pt}
\resizebox{0.999\linewidth}{!}{
\begin{tabular}{c|cc|cc | cc}
\specialrule{.15em}{.05em}{.05em}
Method & 
\multicolumn{2}{c|}{
CBSD68~\cite{Martin_Fowlkes_Tal_Malik_2001}} 
& \multicolumn{2}{c|}{Kodak24~\cite{Franzen_1999}} 
& \multicolumn{2}{c}{McMaster~\cite{Zhang_Wu_Buades_Li_2011}}\\
Noise-level $\sigma$ & 25 & 50 & 25 & 50 & 25 & 50\\
\hline
DnCNN~\cite{Zhang_Zuo_Chen_Meng_Zhang_2017} 
 & 31.24 & 27.95
 & 32.14 & 28.95
 & 31.52 & 28.62\\
DnCNNTV 
 & 31.26 & \bf 28.07 
 & 32.15 & \bf 29.09
 & \bf 32.32 & \bf 29.35\\ 
\specialrule{.15em}{.05em}{.05em}
\end{tabular}
}
\vspace{-0.1cm}
\caption{Quantitative comparison for image denoising. We report the average PSNR ($\uparrow$) for each of the methods.}
\label{tab:denoise_results}
\vspace{-0.2cm}
\end{table}

%% file: sec_conc.tex
\section{Conclusion}\label{sec:conc}
Optimization as a layer is a promising direction to incorporate inductive bias into deep-nets. In this work, we propose to include total variation minimization as a layer. Our method achieves $37\times$ speedup over existing solutions scaling TV layers to real computer vision tasks. On five tasks, we demonstrate existing deep-net architectures can benefit from use of TV-layers. We believe the TV layer is an important building block for deep learning in computer vision and foresee more applications to benefit from it.

\noindent\textbf{Acknowledgements:} This work is supported in part by NSF \#1718221, 2008387, 2045586, 2106825, MRI \#1725729, NIFA 2020-67021-32799 and Cisco Systems Inc.\ (CG 1377144 - thanks for access to Arcetri).